\documentclass[conference]{IEEEtran}
\IEEEoverridecommandlockouts
\usepackage{cite}
\usepackage[utf8]{inputenc}
\usepackage{comment}
\usepackage{array}
\usepackage{longtable}
\usepackage{xspace} 
\usepackage{color}
\usepackage{amsmath}
\usepackage{graphicx}
\usepackage[table]{xcolor}
\usepackage{verbatim}
\usepackage{hyperref}
\def\BibTeX{{\rm B\kern-.05em{\sc i\kern-.025em b}\kern-.08em
    T\kern-.1667em\lower.7ex\hbox{E}\kern-.125emX}}

\begin{document}

\title{Playing Carcassonne with Monte Carlo Tree Search
}

\author{\IEEEauthorblockN{ Fred Valdez Ameneyro$^*$\thanks{$^*$Joint first authors}}
\IEEEauthorblockA{{Naturally Inspired Computation}\\{Research Group} \\
{Department of Computer Science}\\ {Hamilton Institute}\\
Maynooth University, Ireland \\
fred.valdezameneyro.2019@mumail.ie}
\and
\IEEEauthorblockN{ Edgar Galván$^{*,+}$\thanks{$^+$Senior Author}}
\IEEEauthorblockA{{Naturally Inspired Computation}\\{Research Group} \\
{Department of Computer Science}\\ {Hamilton Institute}\\
Maynooth University, Ireland \\
edgar.galvan@mu.ie}
\and
\IEEEauthorblockN{Ángel Fernando\\ Kuri Morales}
\IEEEauthorblockA{{Department of Computer Science} \\
{National Autonomous}\\{ University of Mexico}\\
Mexico city, Mexico \\
akuri@itam.mx}
}

\IEEEoverridecommandlockouts
\IEEEpubid{\makebox[\columnwidth]
{978-1-7281-2547-3/20/\$31.00~\copyright2020 IEEE \hfill} 
\hspace{\columnsep}\makebox[\columnwidth]{ }}
\IEEEpubidadjcol

\maketitle

\begin{abstract}
Monte Carlo Tree Search (MCTS) is a relatively new sampling method with multiple variants in the literature. They can be applied to a wide variety of challenging domains including board games, video games, and energy-based problems to mention a few. In this work,
we explore the use of the vanilla MCTS and the MCTS with Rapid Action Value Estimation (MCTS-RAVE) in the game of Carcassonne, a stochastic game with a deceptive scoring system where limited research has been conducted. We compare the strengths of the MCTS-based methods with the Star2.5 algorithm, previously reported to yield competitive results in the game of Carcassonne when a domain-specific heuristic is used to evaluate the game states. We analyse the particularities of the strategies adopted by the algorithms when they share a common reward system. The MCTS-based methods consistently outperformed the Star2.5 algorithm given their ability to find and follow long-term strategies, with the vanilla MCTS exhibiting a more robust game-play than the MCTS-RAVE.


\end{abstract}


\section{Introduction}

Monte Carlo Tree Search (MCTS) is a sampling method for finding \textit{optimal decisions} by performing random samples in the decision space and building a tree according to partial results.  In a nutshell, Monte Carlo methods work by approximating future rewards that can be achieved through random samples. The evaluation function of MCTS relies directly on the outcomes of simulations. Thus, the accuracy of this function increases by adding more simulations. The optimal search tree is guaranteed to be found with infinite memory and computation~\cite{kocsis2006bandit}. However, in more realistic scenarios (e.g., limited computer power), MCTS can produce very good approximate solutions.

MCTS has gained popularity in two-player board games partly thanks to its recent success in the game of Go~\cite{alphago}, which include beating professional human players. The space of solutions ($10^{170}$) and the large branching factor (up to 361 legal moves per turn for each player) makes the game of Go a highly difficult problem for Artificial Intelligence (AI), considered much harder than Chess. The diversification of MCTS in other research areas is extensive. For instance, MCTS has been explored in energy-based problems~\cite{galvan2014heuristic,Galvan_EnergyCon_2014} and in the design of deep neural network (DNN) architectures~\cite{wang2019alphax} (an active research area in evolutionary algorithms (EAs), see~\cite{galvan2020neuroevolution} for a recent comprehensive literature review of EAs in DNNs). The use of MCTS in different research areas, as well as the use of its mechanics employed in other methods~\cite{galvan2020statistical}, can give a good idea of the success of MCTS on challenging problems. Other relevant methods for game playing are EAs, for example in Ms. PacMan~\cite{galvan2010evolving,galvan2010comparing} and board games such as Sudoku~\cite{galvan2009effects,galvan2007towards}.

The versatility and wide applicability of MCTS are due to its different variants \cite{browne2012survey}. For example, a Parallel MCTS~\cite{chaslot2008parallelmcts} has been proposed to take advantage of modern multi-core processors. A single-player based MCTS has also been proposed in board games where no opponent is necessary (e.g., SameGame) to estimate potential winning positions (actions)~\cite{schadd2008spmcts}. Other variants of the algorithm include the modification of some of the core components of MCTS (these components are presented in Section~\ref{sec:background})  specifically in trying to improve its selection mechanism~\cite{coquelin2007bandit}. 

The goal of this work is to compare the particularities of the strategies adopted by the vanilla MCTS, MCTS-RAVE and Star2.5 when they share a common reward system, explained in Section \ref{method}. The Star2.5 algorithm with a hand-crafted heuristic has been reported to perform better in the game of Carcassonne when compared to the MCTS with limited simulations~\cite{heyden2009implementing}. The game of Carcassonne has a relatively high state-space and game-tree complexities, estimated to be $5\cdot10^{40}$ and $8.8\cdot10^{194}$, respectively \cite{heyden2009implementing}. Furthermore, an element that makes this game complex to play using MCTS or indeed any other AI method, is the fact that the scores in Carcassonne can be deceptive, resulting in the algorithms being misled to non-optimal solutions.

The structure of this paper is as follows. Section \ref{sec:background} outlines the background of the game of Carcassonne, Monte Carlo Tree Search (MCTS), MCTS with Rapid Action Value Estimation, and *-minimax. Section \ref{method} presents the  proposed approach. Section~\ref{sec:experimental} explains the experimental setup used and Section \ref{sec:analysis} illustrates and discusses the results obtained by the various methods used in this study, including minimax and MCTS-based methods. The final section offers concluding remarks.


\section{Background}
\label{sec:background}
\subsection{The game of Carcassonne} 
\label{Carcassone}

In this work, the competitive-based version of the game of Carcassonne for two players is considered. This version is a turn-based, perfect information, stochastic, adversarial game with high complexity and wide strategic opportunities. 

\begin{figure}[tb]
\centerline{\includegraphics[scale=0.65]{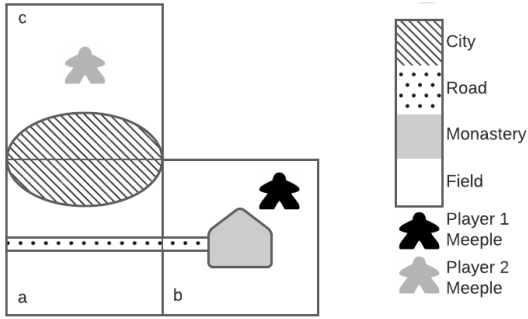}}
\caption{Example of the board of the game of Carcassonne after 2 turns of play. The starting tile ($a$) is the same for each game and is on the board when the game begins. Player 1 played tile $b$ and placed a meeple on the field. Player 2 played tile $c$ with a meeple also on the field. The city in tile $a$ was completed when tile $c$ was played because it cannot be expanded any further. The road in tiles $a$ and $b$ is still incomplete.}
\label{fig_carcassonne_board}
\end{figure}

Carcassonne is a game where players take alternate turns intending to score more points than the opponent. To score points, the players use figures called meeples. The game begins with the starting tile on the board and a stack of 71 tiles to randomly draw from. The board grows after each turn as the players add tiles to it and the game finishes when the tiles are exhausted or if there are no legal actions for the tiles that are remaining. One example of the board of a game of Carcassonne is shown in Figure~\ref{fig_carcassonne_board}. 

Each player's turn consists of three phases:
\begin{itemize}
\item \textit{Placing a tile:} A tile is drawn from the stack of tiles at random. Then, the player chooses any valid spot on the board to play the tile (if there are no legal actions available, the tile is returned to the stack of tiles and a new random tile is drawn). Tile placement follows two rules: (i) the tile must be played in contact with at least one tile that is already on the board, and (ii) all the features on the edges of the played tile that are in contact with other tiles must match the features on the edges of those tiles. Features can be roads, cities, monasteries, or fields, each one with their unique scoring and completion rules. 
\item \textit{Placing a meeple:} The player can choose whether or not to play a meeple in a feature in the played tile unless that feature already has a meeple in it.
\item \textit{Scoring a feature:} All the features with meeples that were completed with the tile placement are scored and the meeples in them are returned to their owners.
\end{itemize}

The meeples are the only way to score points and are limited to 7 tiles for each player. If a meeple is played, it remains on the board and cannot be used again until the related feature is completed. A completed feature gives points to the player with more meeples in it or to both players if their meeples in the feature are the same. 

When the game finishes, all the incomplete features with meeples in them are scored and the player with the most points is declared the winner of the game. The \textit{virtual scores} can be calculated at any point in the game by scoring all the incomplete features on the board \textit{as-if-the-game-is-finished}. Virtual scores and the final scores are the same in the last turn of the game. The virtual scores are more useful than the raw scores to judge the state of the game and are the ones that are going to be used to evaluate game-states in this work, which is explained with deeper detail in Section~\ref{method}.

\subsection{Monte Carlo Tree Search}
\label{sec:mcts}

Monte Carlo Tree Search (MCTS) \cite{kocsis2006bandit} is a sequentially best-first tree search algorithm that grows a tree as it collects data from the state space using Monte Carlo simulations. MCTS iterates four steps and converges to a full tree with minimax values at a polynomial rate if allowed to run indefinitely. The MCTS returns the most promising action from the current state according to a \textit{recommendation policy}. In the \textit{selection} step of the MCTS, the algorithm chooses the best child node at each level of the tree according to a \textit{tree policy} starting from the root node until a leaf node is reached, referred to as the selected node. The \textit{expansion} step adds an unexplored child of the selected node to the tree. The newly added node becomes the expanded node. The \textit{simulation} step simulates entire sequences of actions from the expanded node until a final state is reached. The nodes visited during the simulation step are not stored in the tree and are chosen following a \textit{default policy}. The rewards collected from the simulations are used in the \textit{backpropagation} step to update the statistics of all the nodes that connect the expanded node to the root. The updated statistics help the tree policy to make more informed decisions in the upcoming selection steps, allowing the MCTS to converge to an optimal decision from the root as more iterations are performed. 
 
MCTS uses the Upper Confidence Bounds (UCB1) policy \cite{auer2002finite} as the tree policy (UCT) \cite{coulom2006efficient}, which behaves \textit{optimistically in the face of uncertainty}. The UCT policy is attracted to the less explored nodes because their values are relatively uncertain and it is optimistic about their potential. The parameter $C$ is a scalar constant that balances exploration and exploitation in the UCT function (Eq. \ref{eq_UCT}),

\begin{equation}
\text{UCT} = \frac{r_j}{n_j} + C \sqrt{\frac{ln(n_i)}{n_j}}
\label{eq_UCT}
\end{equation}

The tree stores the number of visits $n_j$ and the reward $r_j$ of each node $j$, where $j\neq root\:node$ and $i$ is the parent of node $j$. The tree policy in the vanilla MCTS selects the child node with the highest UCT value.

\subsection{Monte Carlo Tree Search with Rapid Action Value Estimation}

The Rapid Action Value Estimation (RAVE) \cite{gelly2007mctsrave} is an \textit{all-moves-as-first} (AMAF) heuristic that assumes that there will be a similar outcome from an action regardless of when it is performed. It has been applied to several games, such as the game of Go \cite{gelly2011mctsravego} \cite{rimmel2010biasing} \cite{tom2010computational}, havannah \cite{rimmel2010biasing}, hex \cite{huang2013mohexmctsrave}, and General Video Game Playing \cite{frydenberg2015mctsravegvgp}, to mention some examples.

RAVE works as follows; let $S$ be the list of nodes that share node $i$ as a parent with node $j$, and let $a$ be the action that led to node $j$ from the parent node $i$. In the MCTS-RAVE, the UCT (Eq. \ref{eq_UCT}) is updated to include the AMAF value as shown in Eq. \ref{eq_UCT_RAVE} via~\cite{gelly2011mctsravego}, 

\begin{equation}
\text{UCT}_{RAVE} = (1-\beta_j)\cdot \frac{r_j}{n_j} + \beta_j \cdot \frac{\tilde{r}_j}{\tilde{n}_j} + C \sqrt{\frac{log(n_i)}{n_j}}
\label{eq_UCT_RAVE}
\end{equation}

\noindent where $\tilde{r}_j$ is the accumulated reward from the total of $\tilde{n_j}$ simulations performed from all the nodes in $S$ that included the action $a$ and $\beta$ is a weighting parameter.

The parameter $\beta_j$ is the {minimum MSE schedule}~\cite{gelly2011mctsravego} that minimises the mean squared error between the UCT's reward and the reward including the AMAF values. The minimum MSE schedule has a empiric parameter $b$ called {RAVE bias} and is defined in Eq.~\ref{eq_RAVE_bias},

\begin{equation}
\beta_j = \frac{\tilde{n}_j}{n_j+\tilde{n}_j+4n_j\tilde{n}_jb^2}
\label{eq_RAVE_bias}
\end{equation}

The $\text{UCT}_{RAVE}$ with the minimum MSE schedule uses the AMAF values to greater influence the search when a few simulations have been performed and switches back to the normal UCT behaviour as there is more confidence in the expected rewards. 

\subsection{*-minimax}

The classic \textit{minimax} search is expanded for stochastic games as the \textit{expectimax}  algorithm \cite{ballard1983minimax} \cite{hauk2004m}. Expectimax handles chance nodes by weighting their minimax values according to the probabilities of the respective events. The *-minimax family of algorithms, including  Star1, Star2, and Star2.5, are expectimax variants that use an alpha-beta pruning technique adapted for stochastic trees.

\begin{figure}[tb]

\centerline{\includegraphics[scale=0.28]{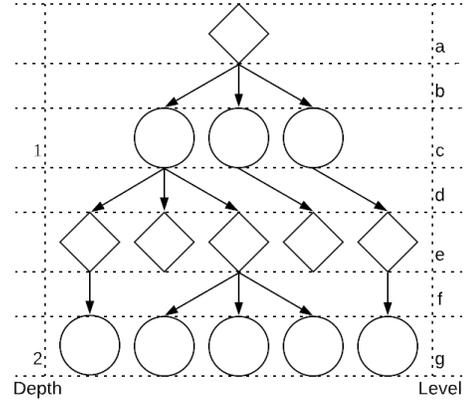}}
\caption{Example of a regular *-minimax game-tree. The root node in level $a$ is the game-state right after a random event happened (rolling the dice in Backgammon or drawing a tile in Carcassonne) and a decision is to be made from there. Edges at levels $b$ and $f$ are actions, and edges at level $d$ are the stochastic events. The depth of the tree reflects the number of decisions that are needed from either player to reach the corresponding game-states. Diamond-shaped nodes are called chance nodes.}
\label{fig_regular_stochastic_tree}
\end{figure}

 In the Star1 algorithm, the theoretical maximum value $U$ and the theoretical minimum value $L$ are used as the guess for the worst and best scenarios of the chance nodes that have not been evaluated in an attempt to prune the tree if the predicted values fall outside an $\alpha\beta$ window as in alpha-beta pruning. In the worst-case scenario, no nodes are pruned and the search behaves as the normal expectimax.

 Star2 is meant for \textit{regular *-minimax games}, in which the actions for each player are influenced by a stochastic event at the beginning of each turn. Examples of regular *-minimax games are Backgammon, Catan, Monopoly, and Carcassonne. A regular *-minimax game-tree is shown in Figure~\ref{fig_regular_stochastic_tree}. In Star2, the first node is evaluated and used as a guessing value for the rest of the sister nodes to prune, in the {probing phase} as in Star1. Thus, ordering of the actions is required to get more reliable results and to prune more often, leading to a faster computational calculation. The actions available from each state are ordered as soon as each state is reached for the first time according to how promising they are (best to worst). The ordering is done following a heuristic that is cheaper than the simulation of the action and the evaluation of the resulting state. If the probing phase fails to achieve a cut-off, the search behaves as the Star1.

A \textit{probing factor} $f>1$ can be predefined in the Star2.5 algorithm. The probing factor determines the number of nodes to be evaluated during the probing phase. In other words, $f=0$ stands for Star1, $f=1$ refers to Star2 and $f>1$ is the Star2.5 method. The Star2.5 method prunes regular *-minimax trees by evaluating the first $f>1$ children, referred to as the probed nodes, of each of the chance events from a node during the probing phase. The algorithm assigns the minimum (in case of a max node) or the maximum (in case of a min node) found in the probed nodes to each of their not-evaluated siblings, which is a safe bet assuming that the nodes are ordered, meaning that no minimum or maximum values are hidden in the not-evaluated nodes. The evaluations are multiplied by the probability of each of their corresponding events to get a weighted sum that is compared to the $\alpha\beta$ window. If the value falls inside of the window, all the children nodes are evaluated to get an accurate value, otherwise, the sub-tree can be pruned. To the best of our knowledge, the sequential Star2.5 algorithm from \cite{ballard1983minimax} is state-of-the-art in the game of Carcassonne using a heuristic to evaluate each node instead of the raw scores or the virtual scores \cite{heyden2009implementing}.  


\section{Proposed Approach} \label{method}

In Carcassonne, the player who plays first has an advantage. Thus, to be fair, the agents are faced against each other in matches of $g$ games, where each agent plays as the first player in half of the games and as the second player in the second half of the games. To guarantee equality of circumstances, the stochastic events from the first half of the match are repeated in the second half.

The goal of each agent is to maximise the reward $r_{p_1}$, Eq.~\ref{eq_reward}, which is proportional to the winning chances in the game of Carcassonne. The reward $r_{p_1}$ of the agent $p_1$ is the difference between the virtual scores $v_{p_1}$ and $v_{p_2}$, where $p_2$ refers to the opponent's score. Note that the virtual scores and the final scores are the same when the game is finished.

\begin{equation} 
\label{eq_reward}
r_{p_1}=v_{p_1,i} - v_{p_2,i}
\end{equation}

Using $r_{p_1}$ as the reward is more consistent than directly maximising the virtual score or returning the outcome of the game (1 for a win, 0 for a draw, and -1 for a loss). To illustrate this point, a list of hypothetical game results are presented in Table \ref{tab:hypothetical_game_results}. For example, despite the agent attaining a higher final score in Game 2 compared to its opponent, it cannot be concluded that the agent performed better than its opponent in Game 1. The agent also performed worse in Game 3 than in Game 1, showing that the final scores can be misleading. One can argue that naively maximising the final score would increase the winning chances, but this is not the case in Carcassonne. The game of Carcassonne allows challenging strategies to emerge where the player focuses his attention on hindering the opponent's scoring chances. For instance, if a player has the opportunity to score the same amount of points with two different actions, the action that interrupts the opposing player's strategy can be the best option. Using the outcome instead of the reward $r_{p1}$ would make the agent blind to opportunities to increase or reduce the gap between the scores, which is desired to ensure winning a game with an advantage or to fight for a come back when losing. 

All the algorithms used in this work, including Vanilla MCTS, MCTS-RAVE, and Star2.5, use Eq.~\ref{eq_reward} to evaluate game-states.

\begin{table}[tb]
\caption{Hypothetical game results}
\begin{center}
\begin{tabular}{|c|c|c|c|c|}
\hline
Game
& \shortstack{Final score\\of the agent}
& \shortstack{Final score of the\\agent's opponent}
& $r_{p1}$
& Outcome\\
\hline
1&70&40&30&1\\
\hline
2&120&90&30&1\\
\hline
3&80&80&0&0\\
\hline
\end{tabular}
\label{tab:hypothetical_game_results}
\end{center}
\end{table}


\section{Experimental setup}
\label{sec:experimental}
The stochastic nature of Carcassonne only influences the order in which the tiles are drawn by the players. A set of 100 sequences of tiles were predefined at random, where each sequence represents the order of the 71 tiles in the stack of tiles for a game. If a match consists of $g$ games, the first half and the second half of the match ($\frac{g}{2}$ games each) is played with the same $\frac{g}{2}$ predefined stacks of tiles. In this way, the stochastic events from each game with the agent $p$ as the first player (first half of the match) are repeated in the games where the agent $p$ is playing as second (second half of the match). When a tile that cannot be played is drawn by a player, that tile is returned to the bottom of the stack and a new tile is drawn from the top of the stack, this occurred in approximately 2.3\% of the total number of games in all the experiments carried out in this work. The sequence of the tile stack is restored after that game is finished.

All MCTS-based agents returned the child node with the most visits as their recommendation policy. The value $r_{p_1}$ (Eq. \ref{eq_reward}) is returned as the reward in the MCTS simulations. When more than one simulation is performed in the same simulation step, the mean value $\bar{r}_{p_1}$ of $r_{p_1}$ is returned. It is possible to return the reward from Eq. \ref{eq_reward} or to use a heuristic function instead of simulating entire games to evaluate any state that is not a final state, which is proposed as future work. Some of the benefits of the Monte Carlo simulations are sensitivity to long-term strategies and robustness against potential deceptive scenarios as discussed previously, which would need to be reflected in a well-crafted heuristic function.

A random agent (RA) was used for parameter tuning. On its turn, the RA chooses the location to play the drawn tile uniformly at random, then it decides between all the available meeple placements or not playing a meeple at all with uniform probabilities. The default policy of the MCTS-based agents makes decisions similar to the RA.

The exploration parameter $C$ was set to 3 as in \cite{heyden2009implementing}. The number of simulations $s$ (also called rollouts) per simulation step for the vanilla MCTS agent is set to 100. For the MCTS-RAVE agent, the best parameters were $s=100$ and the RAVE bias $b=10$ from Eq. ~\ref{eq_RAVE_bias}. These values were obtained with preliminary experiments against the RA.

Star2.5 uses the following move order:

\begin{itemize}
\item 1$^{st}$: All actions that place a meeple in a city.
\item 2$^{nd}$: All actions that place a meeple in a monastery.

\item 3$^{rd}$: All actions that place a meeple in a road.

\item 4$^{th}$: All actions with no meeple placement.

\item 5$^{th}$: All actions that place a meeple in a field.
\end{itemize}

 The rest of the parameters for the Star2.5 algorithm are shown in  Table \ref{tab:Star25_parameters}. The move order and the parameters of the Star2.5 agent were inspired by the work carried out in~\cite{heyden2009implementing}. The game-states are evaluated with the reward shown in Eq.~\ref{eq_reward} instead of the hand-crafted heuristic function suggested in \cite{heyden2009implementing} for the game of Carcassonne. The time that the Star2.5 agent takes for each decision is highly sensitive to the branching factor. A depth of 3 was the maximum that allowed the Star2.5 agent to return an action. After some testing, we found out that setting the depth to 4 or 2, resulted in never getting a result due to the computational power required (depth 4) as well as yielding poor performance (depth 2). The Star2.5 algorithm was allowed to run for as long as it needed to search in the tree. Due to the nature of the MCTS, the depth is not a constraint. Thus, we let the MCTS-based algorithms to make a move in relatively the same amount of time. A summary of all the experiments with their associated execution times is shown in Table \ref{tab:Experiments}. To execute this large number of experiments, we used a computer with 336 nodes operated by Irish Centre for High-End Computing. 

\begin{table}[tb]
\caption{Star2.5 parameters}
\begin{center}
\begin{tabular}{|l|l|r|}
\hline
Symbol
& Definition
& Value \\
\hline
$b$
& Probing factor
& 5\\
\hline
$L$
& Theoretical minimum value
& -100\\
\hline
$U$
& Theoretical maximum value
& 100\\
\hline
$d_{max}$
& Max depth
& 3\\
\hline
\end{tabular}
\label{tab:Star25_parameters}
\end{center}

\caption{Summary of experiments with their associated execution times.}
\begin{center}
\begin{tabular}{|c|c|c|c|}
\hline
Experiment
& \shortstack{Total \\games}
& \shortstack{Time per \\game (min)}
& \shortstack{Total\\time (min)}\\
\hline
\shortstack{MCTS \textit{vs} RA\\(parameter tuning)}
& 300
& 36
& 10800
\\
\hline
\shortstack{MCTS-RAVE \textit{vs} RA\\(parameter tuning)}
& 900
& 31
& 32400\\
\hline
\shortstack{Star2.5 \textit{vs} RA}
& 100
& 20.58
& 2058\\
\hline
\shortstack{MCTS \textit{vs} Star2.5 }
& 140
& 55.79
& 7811 \\
\hline
\shortstack{MCTS-RAVE \\\textit{vs} Star2.5 }
& 140
& 55.79
& 7811 \\
\hline
\shortstack{MCTS-RAVE \\\textit{vs} MCTS }
& 200
& 71
& 14200\\
\hline
Total
& 1780
& N/A
& 75080\\
\hline
\end{tabular}
\label{tab:Experiments}
\end{center}
\end{table}


\section{Analysis and results}
\label{sec:analysis}

We provide here a discussion of the results, including winning rates, obtained by the proposed approach using MCTS, MCTS with RAVE, and Star2.5. Other relevant domain-specific measurements, not known by the agents, are also quantified and reported in this section:
\begin{itemize}
\item \textit{Meeples per game (mpg)}: The meeples that the agent plays throughout the whole game is expected to be maximised. Playing meeples is the only way to score in the game and are a limited resource that is desired to be exploited and recycled as much as possible. The game rewards the most when features are completed, and a greater feature completion rate would mean more meeples becoming available again.

\item \textit{Turns with meeples (twm)}: The rate of turns in which the player has at least one meeple available to play is expected to be maximised to 1. We propose this metric based on two objectives: maximising immediate-scoring opportunities and ensuring the ability to compete for valuable features. Immediate-scoring opportunities are those where a meeple is played and collected in the same turn if the related feature is completed, normally giving a few but risk-free points to the owner. Immediate-scoring opportunities are normally available in every turn of the game as long as the player has available meeples to claim them. One can argue that if a feature becomes too valuable, it can be worth using the last meeple to claim it. Such scenarios are more likely to happen as the game is closer to its end. This measurement is proposed as an indicator of awareness of future scoring opportunities. The formula to calculate it for the agent $p1$ is shown in Eq.\ref{eq_twm} where $t$ is the number of turns where $p1$ started with at least one available meeple and the constants 28 and 29 are the total turns in a game of Carcassonne where each player can potentially have no meeples available.
\end{itemize}

\begin{equation} 
\label{eq_twm}
    twm= 
\begin{cases}
    \frac{t}{28},& \textit{\text{if} p1 \text{was first to play}}\\
    \frac{t}{29},& \textit{\text{if} p1 \text{was second to play}}
\end{cases}
\end{equation}

Table~\ref{tab:RA_versus_all} shows the results of the games between the best configurations of the MCTS-based agents and the Star2.5-based agent playing as second against the random agent. The MCTS-based agents achieved similar results among themselves, showing the robustness of our approach for these techniques. It is also noteworthy to observe how these MCTS-based agents achieve a high number of meeples per game compared to Star2.5 explaining the high scores achieved in the reward and virtual scores (shown in the second and third columns, from left to right in Table~\ref{tab:RA_versus_all}). The random agent was always beaten by all the algorithms shown in Table~\ref{tab:RA_versus_all}.

\begin{table}[tb]
\caption{MCTS, MCTS RAVE, and Star2.5 agents \textit{vs.} the random agent (RA). Comparison of the results of 100 games with the RA as the first player against the MCTS-based agents and the Star2.5-based agent. $\bar{r}_{p1}$ and $\bar{v}_{p1}$ are the mean of the $r_{p1}$ and $v_{p1}$ values from Eq. \ref{eq_reward}. $\bar{mpg}$ and $\bar{twm}$ are the mean of the $mpg$ and $twm$ scores as explained in the beginning of this Section}
\begin{center}
\begin{tabular}{|c|c|c|c|c|}
\hline
Agent
& $\bar{r}_{p1}$
& $\bar{v}_{p1}$
& $\bar{mpg}$
& $\bar{twm}$\\
\hline
Vanilla MCTS
& 93.45
& 107.44
& 14.98
& 0.57\\
\hline
MCTS-RAVE
& 95.74
& 109.42
& 15.47
& 0.59\\
\hline
Star2.5
& 69.93
& 85.41
& 12.91
& 0.47\\
\hline
\end{tabular}
\label{tab:RA_versus_all}
\end{center}

\caption{ MCTS \textit{vs.} Star2.5. Comparison of  results of 140 games between the vanilla MCTS-based agent \textit{vs.} the Star2.5-based agent. In 70 games,  vanilla MCTS was the first player and Star2.5 was the first player in the remaining games}
\begin{center}
\begin{tabular}{|c|c|c|c|c|}
\hline
Criteria
& \shortstack{MCTS as \\$1^{st}$ player}
& \shortstack{Star2.5 as\\$1^{st}$ player}
& \shortstack{MCTS as \\$2^{nd}$ player}
& \shortstack{Star2.5 as \\$2^{nd}$ player}\\
\hline
Win rate
& 88.57\%
& 22.86\%
& 77.14\%
& 2.86\%\\
\hline
$\bar{r}_{p1}$
& 20.29
& -16.82
& 16.82
& -20.29\\
\hline
$\bar{v}_{p1}$
& 80.54
& 60.29
& 77.11
& 60.26\\
\hline
$\bar{mpg}$
& 13.10
& 11.63
& 12.72
& 11.03\\
\hline
$\bar{twm}$
& 0.48
& 0.41
& 0.49
& 0.41\\
\hline
\end{tabular}
\label{tab:MCTS_versus_Star2.5}
\end{center}
\end{table}

We now consider scenarios by comparing each of the intelligent agents used in this work, including Star2.5 and the MCTS-based techniques, against each other. Given that the first player in the game of Carcassonne has an advantage over the second player, each match is divided in two sets of games (in the second set, the player who went first in all the games of the first set now goes second). The same sequence of tiles (generated at random) from the first set of experiments is also used in the second set for fairness. Tables  \ref{tab:MCTS_versus_Star2.5}, \ref{tab:MCTS-RAVE_versus_Star2.5} and~\ref{tab:MCTS_versus_MCTS-RAVE} show the win rate, the mean of the difference of the final scores (Eq. \ref{eq_reward}) $\bar{r}_{p1}$, the final score $v_{p1}$, the meeples per game $mpg$ and the turns with available meeples $twn$, as explained in Section \ref{method}, achieved by each agent when competing against each other, starting with MCTS \textit{vs.} Star2.5 (Table~\ref{tab:MCTS_versus_Star2.5}), MCTS-RAVE \textit{vs.} Star2.5 (Table~\ref{tab:MCTS-RAVE_versus_Star2.5}) and MCTS-RAVE \textit{vs.} MCTS (Table~\ref{tab:MCTS_versus_MCTS-RAVE}). Note that a draw is a possible outcome in the game of Carcassonne and is not reflected in the winning rates of the Tables \ref{tab:MCTS_versus_Star2.5}, \ref{tab:MCTS-RAVE_versus_Star2.5} and \ref{tab:MCTS_versus_MCTS-RAVE}. Thus, a few of the winning rates in these tables do not equate to 100\%.

As can be seen in Table \ref{tab:MCTS_versus_Star2.5}, the vanilla MCTS agent can beat the Star2.5 agent consistently: in 88.57\% of the games, MCTS won when playing first and 77.14\% when playing second, shown in the second row from top to bottom in Table~\ref{tab:MCTS_versus_Star2.5}. These results can be explained when one takes a look to the results yielded by MCTS and Star2.5 when considering the rest of the elements computed in this study. For example, the final scores attained by the MCTS-based agents are consistently better against those yield by Star2.5. This is aligned with the results obtained when computing the number of meeples per game. In cases when some of the latter results are close to each other (for example 13.10 and 11.63), when MCTS is the first player and Star2.5 is the opponent, their impact is significant in their final scores, 85.54 \textit{vs.} 60.29 for MCTS and Star2.5, respectively.

\begin{figure}[tb]
\centerline{\includegraphics[scale=0.6]{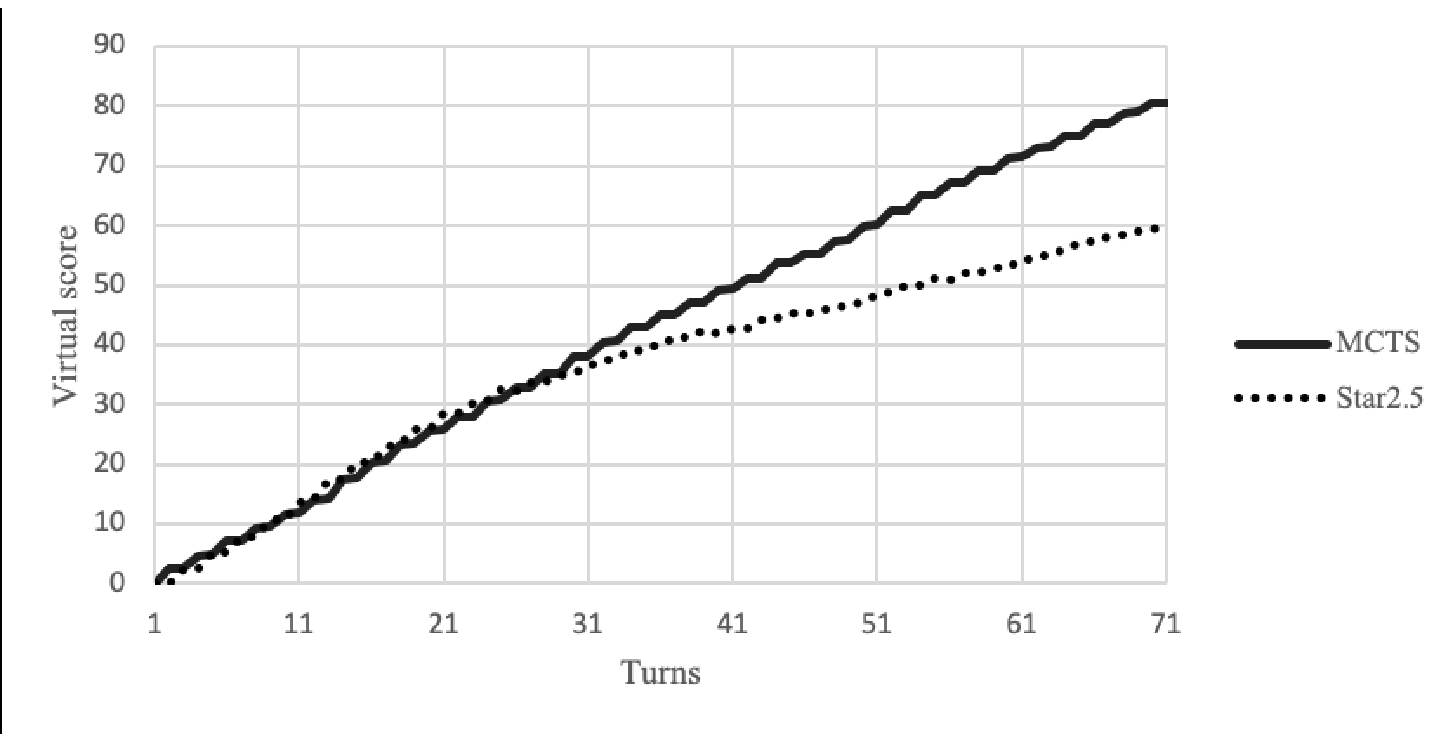}}
\caption{Mean of the virtual scores at each turn of play in the 100 games with MCTS as the first player and Star2.5 as second.}
\label{fig_mcts-s25}

\centerline{\includegraphics[scale=0.6]{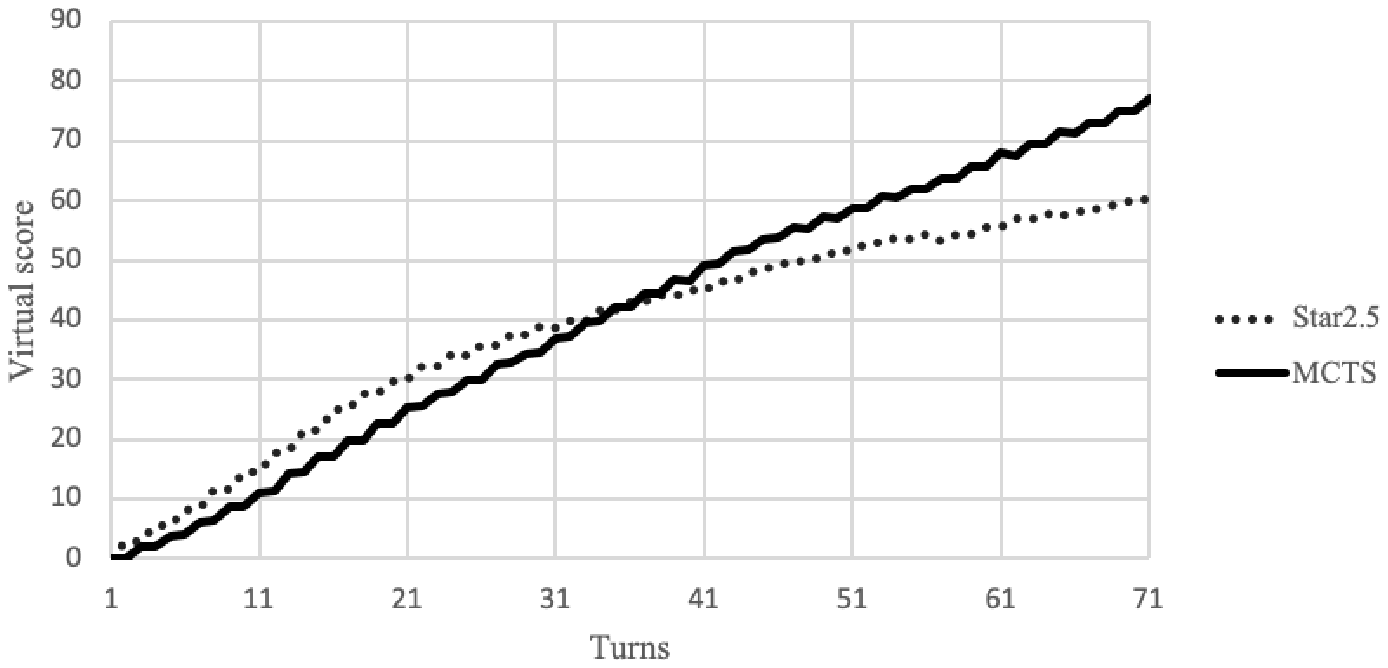}}
\caption{Mean of the virtual scores at each turn of play in the 100 games with Star2.5 as the first player and MCTS as second.}
\label{fig_s25-mcts}
\end{figure}

From this analysis, it is clear that playing first has an advantage in the game of Carcassonne as seen by the results shown in Table~\ref{tab:MCTS_versus_Star2.5}. However, it is unclear how each of these two algorithms behave during the game. To shed some light on this, we keep track of the performance, measured in terms of virtual scores, when MCTS plays first (Figure~\ref{fig_mcts-s25}) and when Star2.5 plays first (Figure~\ref{fig_s25-mcts}). It is interesting to note that MCTS clearly dominates Star2.5 when the former plays first throughout the game. This situation, however, changes when Star2.5 plays first. The MCTS-based agent performs worse compared to Star2.5 for the first half of the game ultimately dominates Star2.5. We hypothesise that the strength of the Star2.5-based agent is to find the best actions that grant the most points in the short term, but it fails in the long term.

From the previous analysis it is clear that MCTS is stronger than Star2.5, an algorithm reported to yield competitive results in the Carcassonne game \cite{heyden2009implementing}. To see the robustness of the approach proposed in this work, we now examine how this behaves when used in MCTS-RAVE and compare the results obtained by this algorithm against Star2.5. Table \ref{tab:MCTS-RAVE_versus_Star2.5} shows these results. It can be observed that MCTS-RAVE consistently beats Star 2.5 regardless of whether it is the first or the second player as seen in the second and fourth column in Table~\ref{tab:MCTS-RAVE_versus_Star2.5}, where  it wins 84.29\% and 78.57\% respectively. This confirms how the agent that plays first has an advantage over its opponent (see, for instance, how Star2.5 achieved a win rate of 20\% and 14.29\%, respectively).

We have seen how the proposed method is consistent when adopted in vanilla MCTS and MCTS-RAVE when each of these is used independently in the game of Carcassonne against Star2.5. We illustrate how these two MCTS-based algorithms behave when competing against each other. The results of these are shown in Table~\ref{tab:MCTS_versus_MCTS-RAVE} for 200 independent games. When MCTS-RAVE plays first, it achieves 51\% win rate and falls short against MCTS when it starts as second player, as shown in the fourth column, from left to right, in Table~\ref{tab:MCTS_versus_MCTS-RAVE}. One would have expected for the MCTS-RAVE to outperform the vanilla MCTS as the former is reported to generally do so in domains with similarities with the game of Carcassonne. An in-depth analysis of the associated games showed that MCTS-RAVE agent behaved greedier than the vanilla MCTS, illustrated by a lower $twm$ score, which is a contrasting result with the $twm$ values from Table \ref{tab:RA_versus_all}, where the vanilla MCTS achieved lower $twm$ than the MCTS-RAVE against the RA. Both algorithms (MCTS and MCTS-RAVE) used their meeples on each feature type in a similar fashion, but not in the farms. The MCTS-RAVE used fewer meeples on farms than the vanilla MCTS on average (2.8 for the MCTS-RAVE $vs$ 3 from the vanilla MCTS). A farm is a peculiar feature in the game of Carcassonne which can never be completed, meaning that any meeple played on it cannot be retrieved again. It is also a feature that can be easily extended with new tiles, or connected with other existing farms, often becoming extensive and very valuable. Early into the game, it is not optimum to use meeples on the farms, but as the game progresses the value of the farms increases and it becomes harder to fight for them if they are already claimed. We hypothesise that the vanilla MCTS is more robust when it comes to assessing the value of the farms and when it is the right time to claim them.

\section{Discussion}

To objectively compare the strengths of intelligent agents created to play games with rich strategical game-play such as the game of Carcassonne is not trivial. The MCTS-RAVE agents showed superior performance than the vanilla MCTS in their games against the RA, but when faced against each other, vanilla MCTS marginally outperformed the MCTS-RAVE, generating seemingly conflicting results. This illustrates the need for a bigger sample of agents to make conclusions about the robustness of the algorithms with more confidence. Robustness is a crucial characteristic for any agent that is to play the game of Carcassonne, and exhaustive experiments against agents with particular behaviours are needed to measure their strength in the future.

Parameter tuning of the MCTS-based agents was performed based on the results from games against a random agent, which given what we stated above, could mean that the agents were optimised to exploit the random agent mistakes instead of developing a robust strategy. It is unlikely that a random agent happens to follow long-term strategies by chance in the game of Carcassonne given its immense state-space. By using the RA for parameter tuning, the MCTS-based agents were unable to optimise to face long-term strategies. The development of multiple agents with particular behaviours in the game of Carcassonne is then seen as a stepping stone to perform proper parameter tuning and objective strength evaluations.

It is worth indicating that our findings are not comparable with those achieved by Heyden \cite{heyden2009implementing} given the differences in the experimental setups; like the stopping criteria for each agent (In \cite{heyden2009implementing} was given 6 minutes for the full game, with no clarification on the number of iterations that the MCTS agent was able to perform), the number of simulations $s$ for the vanilla MCTS ($s=1$ in \cite{heyden2009implementing}, compared to our $s=100$ after parameter tuning), the evaluation of game states in the Star2.5 algorithm (hand-made heuristic in \cite{heyden2009implementing} instead of the reward $r$), and the forced equivalence of random events that we used. The results from Table \ref{tab:MCTS_versus_Star2.5} compare the strengths of the Star2.5 and the vanilla-MCTS agents in an even field when no domain-specific knowledge is used, allowing us to visualise the behaviour, strengths and weaknesses of each algorithm in this particular game. As expected, the limited depth of the Star2.5 algorithm allows it to find good short-term solutions but ignoring long-term consequences.

\begin{table}[tb]
\caption{MCTS-RAVE \textit{vs} Star2.5. Comparison of the results of 140 games between the MCTS-RAVE agent against the Star2.5 agent. 70 games had MCTS-RAVE as the first player and 70 had Star2.5 as the first player. $\bar{r}_{p1}$ and $\bar{v}_{p1}$ are the mean of the $r_{p1}$ and $v_{p1}$ values from Eq. \ref{eq_reward}. $\bar{mpg}$ and $\bar{twm}$ are the mean of the $mpg$ and $twm$ scores as explained in the beginning of this Section.}
\begin{center}
\begin{tabular}{|c|c|c|c|c|}
\hline
Criteria
& \shortstack{MCTS-RAVE \\as $1^{st}$ player}
& \shortstack{Star2.5 \\as $1^{st}$\\player}
& \shortstack{MCTS-RAVE \\as $2^{nd}$ player}
& \shortstack{Star2.5  \\as $2^{nd}$\\player}\\
\hline
Win rate
& 84.29\%
& 20.00\%
& 78.57\%
& 14.29\%\\
\hline
$\bar{r}_{p1}$
& 18.30
& -17.54
& 17.54
& -18.30\\
\hline
$\bar{v}_{p1}$
& 80.66
& 62.46
& 80.00
& 62.36\\
\hline
$\bar{mpg}$
& 13.33
& 11.48
& 13.17
& 11.11\\
\hline
$\bar{twm}$
& 0.5
& 0.41
& 0.52
& 0.4\\
\hline
\multicolumn{5}{l}{}
\end{tabular}
\label{tab:MCTS-RAVE_versus_Star2.5}
\end{center}
\end{table}


\begin{table}[tb]
\caption{MCTS-RAVE \textit{vs} MCTS. Comparison of the results of 200 games between the MCTS-RAVE agent against the vanilla MCTS agent. 100 games had MCTS-RAVE as the first player and 100 had vanilla MCTS as the first player. $\bar{r}_{p1}$ and $\bar{v}_{p1}$ are the mean of the $r_{p1}$ and $v_{p1}$ values from Eq. \ref{eq_reward}. $\bar{mpg}$ and $\bar{twm}$ are the mean of the $mpg$ and $twm$ scores as explained in the beginning of this Section.}
\begin{center}
\begin{tabular}{|c|c|c|c|c|}
\hline
Criteria
& \shortstack{MCTS-RAVE \\as $1^{st}$ player}
& \shortstack{MCTS \\as $1^{st}$ \\player}
& \shortstack{MCTS-RAVE \\as $2^{nd}$ player}
& \shortstack{MCTS \\as $2^{nd}$ \\player}\\
\hline
Win rate
& 51\%
& 64\%
& 36\%
& 49\%\\
\hline
$\bar{r}_{p1}$
& 0.11
& 2.67
& -2.67
& -0.11\\
\hline
$\bar{v}_{p1}$
& 80.05
& 80.75
& 78.08
& 79.94\\
\hline
$\bar{mpg}$
& 13.49
& 13.71
& 12.61
& 13.32\\
\hline
$\bar{twm}$
& 0.5
& 0.51
& 0.48
& 0.51\\
\hline
\end{tabular}
\label{tab:MCTS_versus_MCTS-RAVE}
\end{center}
\end{table}

\section{Conclusions}

We have shown that the MCTS-based algorithms (vanilla MCTS and MCTS with Rapid Action Value Estimation) are capable to find long-term strategies in the game of Carcassonne, a complex and deceptive game. This is an advantage when they are compared with the Star2.5 algorithm when no domain-specific knowledge is involved, and suggest that an MCTS-based agent enhanced with heuristics could be more robust than any Star2.5 agent.

The strength of the agents in the game of Carcassonne can be seen as relative when tested against a limited set of opponents, given the variety of strategies that an agent can adopt to play. This study found that the MCTS-based agents are more robust than the Star2.5 given their ability to plan ahead, but with potential flaws. The MCTS-RAVE and the vanilla MCTS present differences when evaluating plans that involve playing with farms, which marginally favour the vanilla MCTS agent when they are faced against each other. This difference in behaviour favours the MCTS-RAVE when both vanilla MCTS and MCTS-RAVE, are tested against an agent making moves at random, indicating that their strategies are not entirely robust. Despite this, the MCTS-based agents are a promising approach for further research in the game of Carcassonne, showing better characteristics than the Star2.5 algorithm in general. 

Finally, the proposed turns with meeples (twm) metric, designed to measure the awareness of long-term scoring opportunities by the agents, showed to align with the results in all the experiments, and can be of use for heuristic development or as an objective for multi-objective approaches in the game of Carcassonne.

\section*{Acknowledgments}
This publication has emanated from research conducted with the financial support of Science Foundation Ireland under Grant number 18/CRT/6049. The authors wish to acknowledge the DJEI/DES/SFI/HEA Irish Centre for High-End Computing (ICHEC) for the provision of computational facilities and support. The first author thanks CONACyT and UNAM for the financial support received during his postgraduate studies. The authors would like to thank F. Stapleton and D. Healy for their comments.
 
\bibliographystyle{plain}
\bibliography{MyBib.bib}

\end{document}